\documentclass[a4paper, 10pt, conference]{ieeeconf}      
\usepackage{FG2019}

\IEEEoverridecommandlockouts                              
\def\FGPaperID{220} 

\usepackage[nocompress]{cite}

\usepackage[pdftex]{graphicx}
\graphicspath{{figuras/}}
\DeclareGraphicsExtensions{.pdf,.jpeg,.png}

\usepackage{algorithmic}
\usepackage{array}
\usepackage[caption=false,font=footnotesize]{subfig}
\usepackage{url}
\begin{document}
\title{Hand range of motion evaluation for Rheumatoid Arthritis patients} 

\IEEEoverridecommandlockouts\pubid{\makebox[\columnwidth]{978-1-7281-0089-0/19/\$31.00~\copyright{}2019 IEEE \hfill}
  \hspace{\columnsep}\makebox[\columnwidth]{ }}

\author{\parbox{16cm}{\centering
    {\large Luciano Walenty Xavier Cejnog$^1$ and Roberto Marcondes Cesar Jr.$^1$ and Teofilo de Campos$^2$ and Val\'{e}ria Meirelles Carril Elui$^3$}\\
    {\normalsize
    $^1$ Departamento de Ci\^encia da Computa\c{c}\~ao, IME-USP, Rua do Mat\~ao, 1010, S\~ao Paulo, Brazil \\
    $^2$ Departamento de Ci\^encia da Computa\c{c}\~ao, CIC, Universidade de Bras\'{\i}lia, 70910-900, Brazil\\
    $^3$ Departamento de Terapia Ocupacional, FMUSP Ribeir\~ao Preto, Brazil\\
}}
  \thanks{This preprint has been submitted to 14th IEEE International Conference on Automatic Face and Gesture Recognition, FG2019.
    We are grateful to Daniela Goia for her collaboration in the data acquisition. We also thank Janko Calic, Maria da Gra\c{c}a Pimentel, Phillip Krejov and Adrian Hilton for the decisive discussions in the early stages of this project.
    This work received financial support of the S\~ao Paulo Research Foundation (FAPESP), grants \#2014/50769-1, \#2015/22308-2 and \#2016/13791-4. TEdC acknowledges the support of CNPq grant PQ 314154/2018-3.}
}


%


\ifFGfinal
\thispagestyle{empty}
\pagestyle{empty}
\else
\author{Anonymous FG 2019 submission\\ Paper ID \FGPaperID \\}
\pagestyle{plain}
\fi
\maketitle

\begin{abstract}
We introduce a framework for dynamic evaluation of the
fingers movements: flexion, extension, abduction and adduction. This framework estimates angle measurements from joints computed by a hand pose estimation algorithm using a
depth sensor (Realsense SR300). Given depth maps as input, our framework
uses Pose-REN~\cite{chen2017}, which is a state-of-art hand
pose estimation method that estimates 3D hand joint positions using
a deep convolutional neural network. The pose estimation algorithm runs
in real-time, allowing users to visualise 3D skeleton tracking results
at the same time as the depth images are acquired. Once 3D joint poses
are obtained, our framework estimates a plane containing the wrist and
MCP joints and measures flexion/extension and abduction/adduction angles
by applying computational geometry operations with respect to this plane.
We analysed flexion and abduction movement patterns using real data,
extracting the movement trajectories. Our preliminary results show
this method allows an automatic discrimination of hands with Rheumatoid Arthritis (RA) and healthy patients. The angle between joints can be used as an indicative of current movement capabilities and function. Although the measurements can be noisy and less accurate than those obtained statically through goniometry,
the acquisition is much easier, non-invasive and patient-friendly, which shows the potential of our approach.
The system can be used with and without orthosis.  Our framework allows the acquisition of measurements with minimal intervention and significantly reduces the evaluation time. 
\end{abstract}


%
\IEEEpeerreviewmaketitle

\section{Introduction}
\begin{figure}[!t]
\centering
\includegraphics[width=\linewidth]{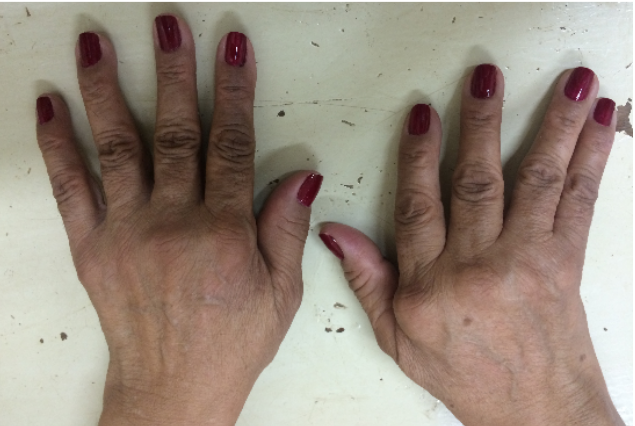}
\caption{Example of hand with finger ulnar deviation (on the right) in contrast with a normal hand (on the left). Such hand shapes represent a challenge for hand tracking algorithms. \ifFGfinal Courtesy of Prof. Valeria Elui. \fi}
\label{fig:hand-patient}
\end{figure}

This paper introduces a computer vision approach to analyze patients evolution on hand occupational therapy. We focus on rheumatoid arthritis (RA) recovery. Rheumatoid arthritis is an autoimmune chronic disease leading to joint deformities. It affects motion functionality of the hand and the treatment requires dynamic and functional evaluations. Typically, \emph{Disabilities of the Arm, Shoulder and Hand} (\emph{DASH}) questionnaires are used to assess hand function during the recovery process and quantitative evaluation uses range of motion measurements.
Figure \ref{fig:hand-patient} shows an example of a hand with ulnar deviation, in contrast with a healthy hand. Our framework should handle both types of hands.  
Although this is a very important health problem, there are few computer vision methods described in the literature to automatically analyze the treatment evolution. This paper represents a step  to fill this gap.

\begin{figure*}[hbt]
\centering
\includegraphics[width=\linewidth]{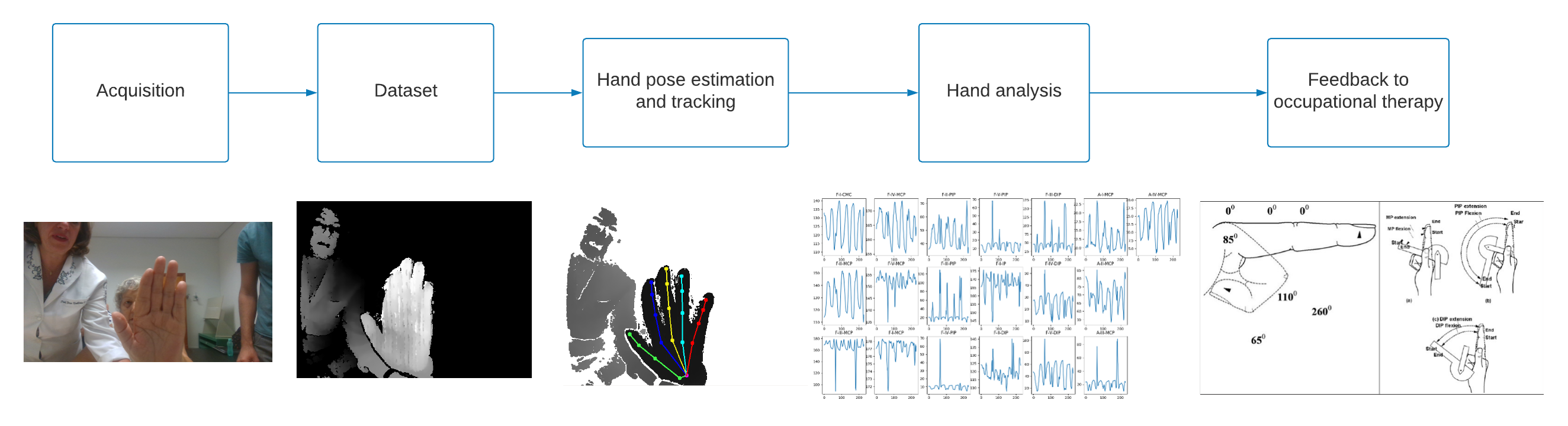}
\caption{Pipeline proposed.}
\label{fig:pipelineProj}
\end{figure*}

The project development follows the pipeline proposed in Figure \ref{fig:pipelineProj}. The pipeline starts with data acquisition (RGB/RGBD), i.e. the acquisition of RGBD sequences from different patients. 
A key step to the proposed approach is to accurately locate hand joints in 3D. Some of the inherent limitations and challenges are the high dimensionality of the hand structure, ambiguities on the model, self-occlusions and abrupt motion. \cite{erol}. 
For each depth image, a state-of-art 3D hand pose estimation method is applied, yielding a skeleton with 21 joints. This skeleton is analyzed in order to estimate range-of-motion measurements, that should be used by the therapist in the treatment. The main contributions of this study are (1) to introduce a new computer vision framework to support hand occupational therapy based on state-of the-art hand tracking;  (2) to introduce the evalutation method based on the estimative of angles and range-of-motion measurements from skeletons; and (3) to describe the preliminary results using real data from patients being treated at the University of S\~ao Paulo hospital.



\section{Related Works}

\subsection{Hand pose estimation}

In recent years, the development of deep learning algorithms led to significant advances in machine learning and its applications, particularly in Computer Vision. The advent of those algorithms combined with the development of accurate solutions for 2D joint detection based on CNNs \cite{wei2016convolutional, newell2016stacked} led the community of hand pose estimation to design methods based on convolutional neural networks\cite{oberweger2017, oberweger,zhou,ge2017, guo2017}.  
Those methods differ among themselves in the neural network architecture and type, the input image type, the hand representation used and the use of prior constraints. As an example, the \textit{DeepPrior++} \cite{oberweger2017} uses a Residual Neural Network, a deep network whose training is based on minimizing residual weights in each layer and uses data augmentation in the training, such that realistic samples can be generated from simple geometric transformations over the original training samples. 
Guo \textit{et. al.}\cite{guo2017} use an ensemble-based neural network, which integrates the results of different regressors in different regions of the image. Chen \textit{et. al.}\cite{chen2017} compute a feature map for each joint and fuse those maps using a structured region ensemble network (named Pose-REN), reaching consistent results. Wan \textit{et. al.}\cite{wan17} propose the use of Generative Adversarial Networks (GAN) and Variational Autoencoder (VAE), two strong ideas in the recent wave of advances in machine learning. 
This method allows training and learning from unlabeled data.

The development of deep learning methods brought the necessity of larger datasets. As a consequence, new million-scale datasets have been made available in 2017: the BigHand2.2M~\cite{yuan2017} and First-Person Action dataset~\cite{garcia2017}. With these datasets, deep learning methods can use a much larger training set and reach better results. To consolidate the trend of using CNNs, the International Conference on Computer Vision board organized the HANDS in the million 2017 challenge on 3D pose estimation~\cite{handsmillion}, a competition on a benchmark using the BigHand2.2M dataset. The results of this challenge were presented in the form of a survey~\cite{yuan20173d}, in which design choices are discussed, as well as the corresponding evaluation results. 

The current panorama of the area indicates that there is room for improvement on methods based on deep CNNs for depth images and that there are efforts of many research groups around the world in this direction. 
This is particularly important in the application addressed in the present paper, since the most state-of-the-art methods trained on standard healthy hands tend to fail in case of hand deformaties such as the one in Figure \ref{fig:hand-patient}.
In parallel, new methods based on learning-based 2D joint detection and Inverse Kinematics are being proposed to estimate hand pose based exclusively on RGB image~\cite{zimmermann, panteleris, mueller2017ganerated, panteleris2017using}.

\subsection{Range of motion measurements}
The evaluation of hand function is fundamental for the therapist to plan the treatment as well as record the results. Literature in hand therapy define metrics and guidelines in order to extract those metrics with precision. For measuring the joint angles a widely used metric is the range of motion (ROM), which consists in a set of angles between joints, whose maximum and minimum values are evaluated during flexion/extension and adduction/abduction movements, usually in a static way, being hard to evaluate during a task performance.
One of the most used assessment methods for range-of-motion measurement is goniometry. With a specific hand/finger goniometer, the therapist can access objectively and reliably the range of motion measurements. The goniometer is widely used due to its simplicity and low cost. However it needs a trained therapist that follows the protocols, is time consuming and really hard to associate these measurements during the execution of activities. If compared to 2D visual estimation and wire tracing, goniometry shows more reliability and precision \cite{ellis1997joint, bruton1999comparison}. 

Digital photogrammetry was used by surgeons, and although its precision is worse if comparable to goniometry, some recent works show that the reliability of this method has increased over the years, being comparable to the ones obtained by a goniometer\cite{carvalho2012analise}. Other alternatives in the evaluation are the use of electronic goniometers, like the torque-based Multielgon system \cite{tajali2016reliability}. 

Among recent works that proposes solutions based on computer vision, Pereira et. al. \cite{pereira2017reliability} proposes a smartphone accelerometer-based app to measure active and passive knee ROM in a clinical setting. The hand case, however, is arguably more challenging than the knee, and 2D hand pose estimation is still not viable. An alternative is the use of depth sensors, and despite its recent rise of popularity few works to date make use of such devices for this task. We highlight the work of Lima \cite{limafisiomotion}, that proposes a system that uses information obtained by a Leap Motion sensor to estimate hand angles. 

We expect that the significant advances in computer vision and hand pose estimation can lead to a series of advances in this specific field of application. The possibility of acquiring 3D frames and skeletons reduces most of ambiguities found in 2D visual estimation, and its use in the treatment of patients can be far less intrusive than the goniometers.


\section{Proposed Approach}

We should follow the pipeline from Figure \ref{fig:pipelineProj}, dividing into data acquisition, 3D hand pose estimation and hand analysis.

\subsection{Data acquisition}
As first step of the project, our goal was to acquire data from patients with hands deformities due to rheumatoid arthritis (AR). Different depth sensors were tested, and among them, the \emph{Intel RealSense\textsuperscript{\textregistered} SR300} generated the best depth maps and has a range which is the most suitable for our acquisition scenario.

Our dataset contains data from both healthy subjects and AR patients. The control group is composed by 10 patients, performing on left and right hands three sequences of flexion and abduction movements, with different movement velocities. For each AR patient and each hand with ulnar deviation, we obtained two flexion and two abdution sequences, with and without the orthosis.

In some of the captured sequences the patient used an orthosis, that is an external mechanical device that applies forces to the body parts (hand joints) in order to enhance the movement capability and function. Quoting \cite{GoiaOrthosis}, "orthoses are external devices applied to any part of the body to stabilize  or immobilize, prevent or correct deformities, protect against injury, maximize function and reduce the pain caused by deformity". In the treatment of fingers ulnar deviation due to rheumatic arthrosis, orthoses are tailor-made by therapists and act like a lever system distributing the force applied to correct the fingers ulnar deviation. 

Table \ref{data-summary} presents a summary of the current state of our dataset.

\begin{table}
\renewcommand{\arraystretch}{1.1}
\centering
\begin{tabular}{|l|r|}
\hline
\multicolumn{2}{|c|}{\textbf{Summary}}\\ \hline
\textbf{Patients with rheumatoid arthritis}   & \textbf{3}\\ 
\textbf{Sequences}  & \textbf{23}\\ 
\textbf{Sequences with orthosis} & \textbf{6}  \\
\textbf{Control sequences with healthy hands} & \textbf{100}\\ 
\textbf{Size (GB)}          & \textbf{147.5}\\ \hline
\end{tabular}
\caption{Summary of our latest dataset.}
\label{data-summary}
\end{table}

\subsection{3D hand pose estimation}

Our current best result for 3D hand pose estimation is Chen et al.'s Pose-REN method \cite{chen2017}, trained with the Hands2017 dataset \cite{yuan20173d}. Pose-REN method is based on the estimation of feature maps using Convolutional Neural Networks (CNNs). Feature maps are combined using an ensemble network, in order to generate a consistent hand pose. 

\begin{figure}[!htb]
    \centering
    \includegraphics[width=0.3\linewidth]{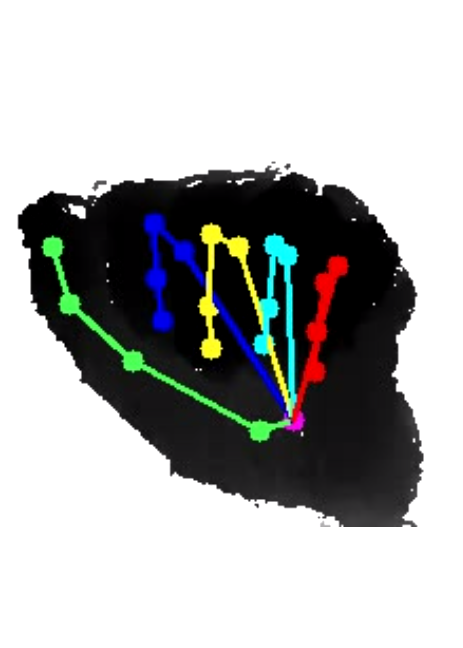}
    \includegraphics[width=0.3\linewidth]{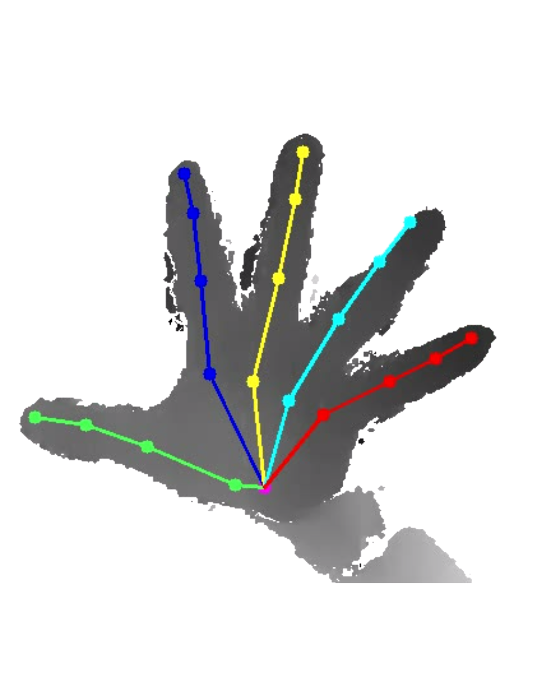}
    \includegraphics[width=0.3\linewidth]{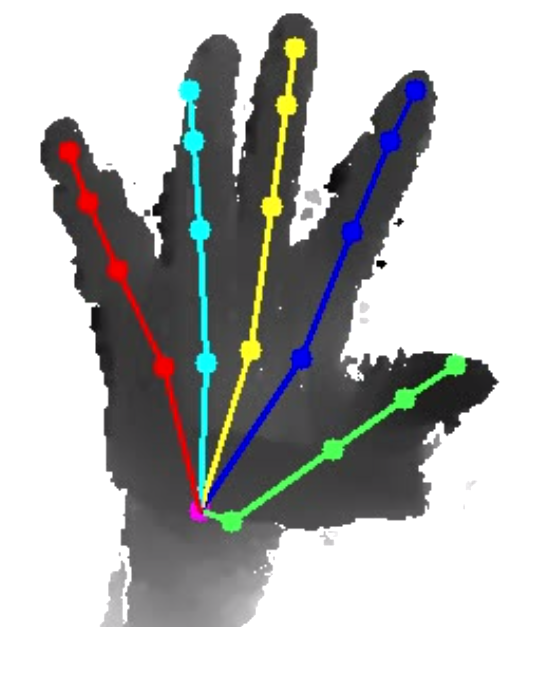}
    \caption{Qualitative results obtained on rheumatoid arthritis patients using the Pose-REN method.}
    \label{fig:guor2}
\end{figure}

This method was chosen due to the ready implementation and the result consistency with the RA patients, as exemplified in Figure \ref{fig:guor2}. 

Even so, it still presents some errors, which indicates that there is still room for improvements as future research (see discussion below).

The skeleton used by Hands17 dataset is composed by 21 joints. Those joints contain the center of the wrist (W) and for each finger x the proximal interphalangeal ($PIP_x$), the distal interphalangeal joints($DIP_x$) and the tip ($TIP_x$). The exception is the thumb, which is represented by the carpometacarpal joint ($CMC$) and a single interphalangeal joint ($IP$). Fingers are represented by the respective roman number (I-V: I for the thumb, V for the pinky finger).

\subsection{Hand analysis}

Using the skeletons obtained in the hand pose estimation method, the analysis aims to obtain relevant measures for our application, i.e. maximum and minimum flexion/extension and adduction/abduction angles.
Our ultimate goal is to estimate these angles with accuracy that is similar to that obtained using manual measurements with goniometers, but in a more efficient and less intrusive way.

The estimation of range of motion angles is divided in flexion and abduction angles, and we geometrically estimate those angles from the skeleton. The estimation of the flexion angles is straightforwardly obtained by extracting the vectors between the adjacent joints in the structure.
\begin{eqnarray}
    &\widehat{FMCP_x} = \arccos{(\overrightarrow{MCP_x - W} \cdot \overrightarrow{PIP_x - MCP_x})}\\
    &\widehat{FPIP_x} = \arccos{(\overrightarrow{PIP_x - MCP_x} \cdot \overrightarrow{DIP_x - PIP_x})}\\
    &\widehat{FDIP_x} = \arccos{(\overrightarrow{DIP_x - PIP_x} \cdot \overrightarrow{TIP_x - DIP_x})}
\end{eqnarray}
For the thumb, the flexion angles of CMC and IP joints are obtained analogously.

The abduction angles represent the ulnar horizontal deviation from the hand in relation to an axis. They are generally measured using the middle finger as reference. For the estimation, we sought to measure the angle between the projection of each finger in the hand palm plane and the vector of the metacarpal bone. For this, we first estimate a plane which represents the hand palm. This is done by computing the normal of the plane that contains $MCP_2$, $MCP_5$ and $W$. Subsequently for each finger $x$ we compute the projection $\vec{P_x}$ of the joint between $PIP_x$ and $MCP_x$ in the plane. Finally the abduction angle is defined as the angle between $\vec{P_x}$ and the joint defined by $MCP_x$ and $W$.
\begin{eqnarray}
    \overrightarrow{N} = \overrightarrow{MCP_2 - W} \times \overrightarrow{MCP_5 - W}\\
    \overrightarrow{P_i} = proj(\overrightarrow{PIP_i - MCP_i}, \overrightarrow{N})\\
    \widehat{AMCP_x} = \arccos{(\overrightarrow{MCP_x - W} \cdot \overrightarrow{P_i})}
\end{eqnarray}

\section{Results}

For all sequences of movement acquired with the patients and with the control individuals we computed the flexion and abduction angles frame by frame. In order to evaluate feasibility of using a hand pose estimation result to access the range of motion, we focused first on flexion movement.

\begin{figure}[!thb]
\centering
\includegraphics[width=\linewidth]{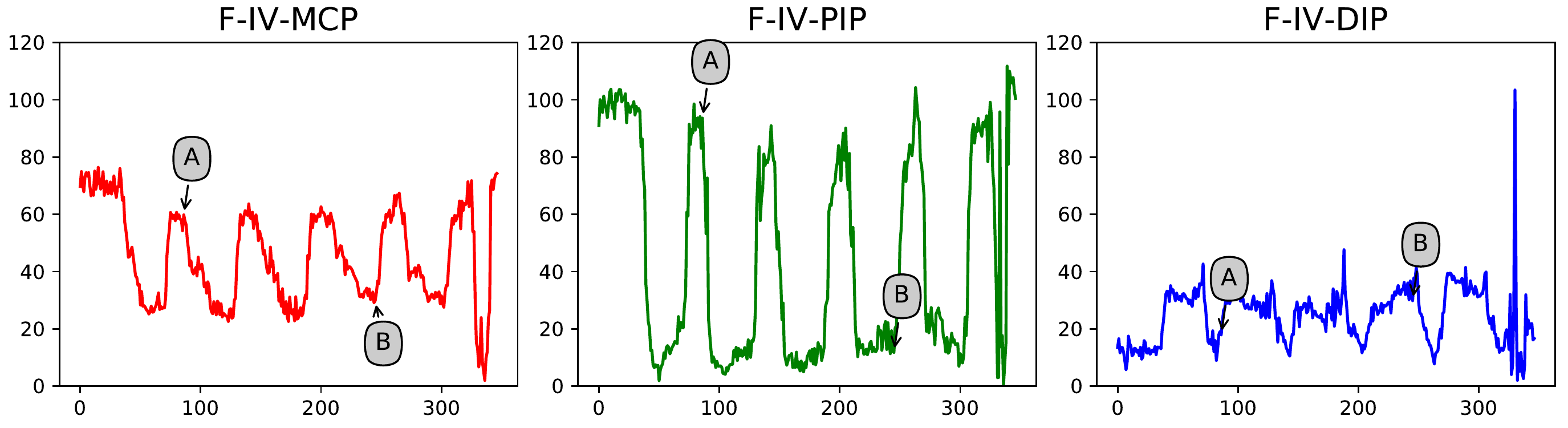}
\caption{Angle evaluation of a patient}
\label{fig:res1}
\end{figure}

\begin{figure}[!thb]
\centering
\includegraphics[width=\linewidth]{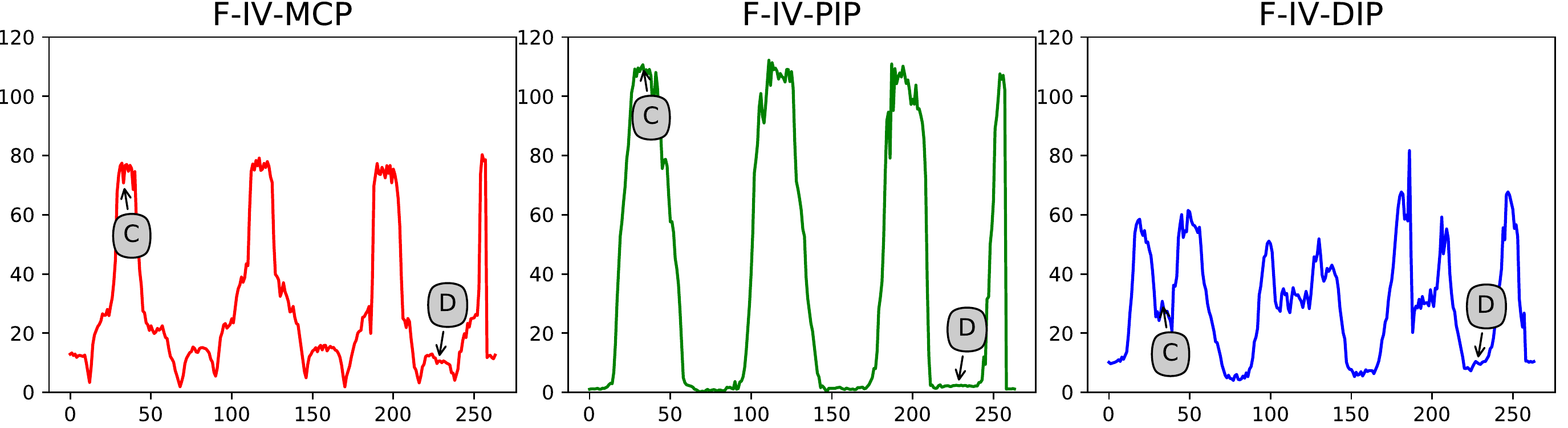}
\caption{Angle evaluation of an individual in control group.}
\label{fig:res2}
\end{figure}

\begin{figure}[!htb]
\centering

\subfloat[Patient - Local maxima A: Hand closed]{\includegraphics[width=0.45\linewidth]{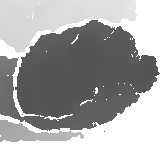}}
\subfloat[Patient - Local minima B: Hand open]{\includegraphics[width=0.45\linewidth]{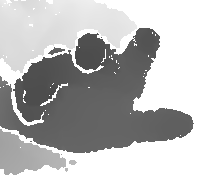}}

\subfloat[Control - Local maxima C: Hand closed]{\includegraphics[width=0.45\linewidth]{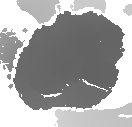}}
\subfloat[Control - Local minima D: Hand open]{\includegraphics[width=0.45\linewidth]{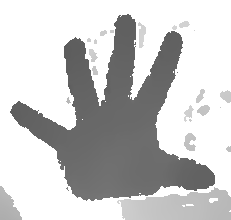}}
\caption{Minima and maxima examples of individuals in control and patient groups.}
\label{fig:res3}
\end{figure}

In the flexion movements, the angle pattern is visible, and the maximum and minimum values of angles translate directly into closed and open hands. Figure \ref{fig:res1} shows the flexion angles for the ring (IV) finger of a patient, while Figure \ref{fig:res2} shows the same angles relative to a member of the control set. Some of the local optima are highlighted in the graph, and Figure \ref{fig:res3} shows the highlighted frames, corresponding to hands closed and open. Thus, each pulse on the graphs correspond to a flexion movement.

Given that the patterns of movement are identified, the next step was the synchronization of the movements. Each sequence contains more than one movement, and we manually identified landmark frames in the beginning and in the end of each flexion movement.

\begin{figure}[htb]
\centering
\includegraphics[width=\linewidth]{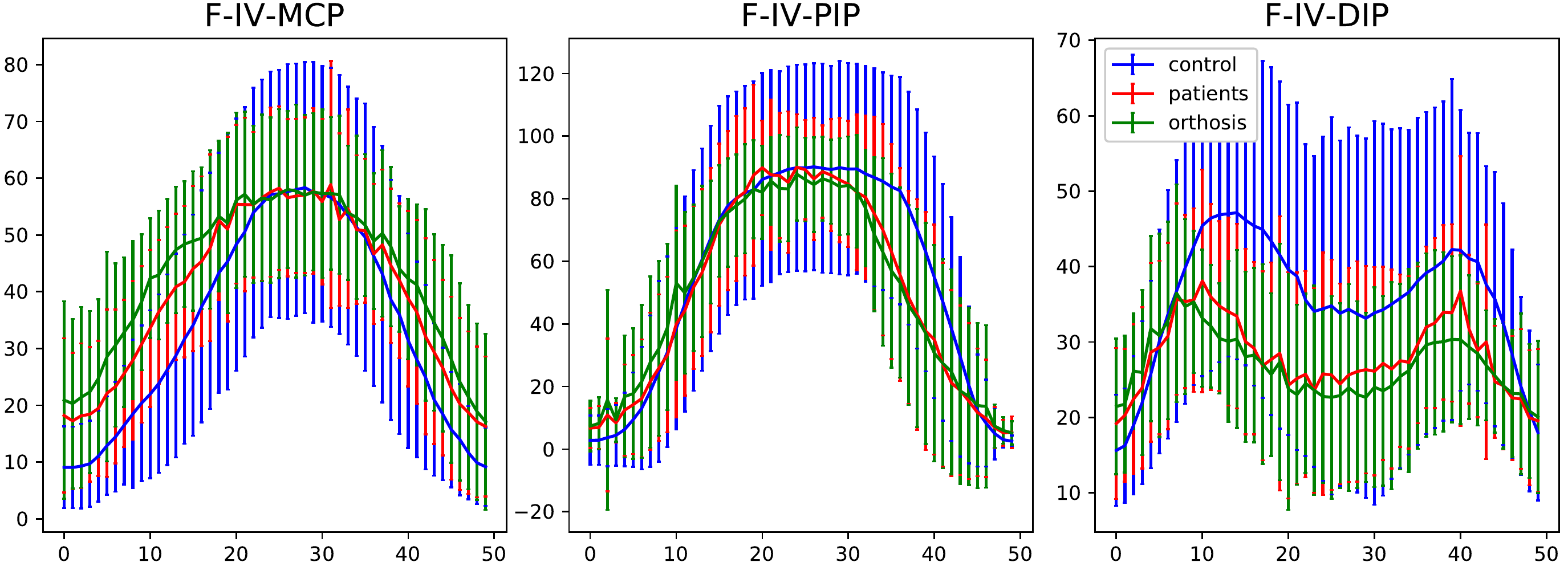}\\
\caption{Comparison of average values and standard deviations obtained in control set (blue), patients with (green) and without orthosis (red) for flexion movement, for finger 4.}
\label{fig:res4}
\end{figure}

With that, we computed the average value and the standard deviation for both patients and control set. This result is shown in Figure \ref{fig:res4}. Note that the graphs have different y-scales. This result shows that both sets follow the same movement pattern and have subtle differences, focused mainly in the beginning of the movement. As illustrated in Figure \ref{fig:res3}, the RA patients have limitations into keeping the hand open widely.

 The method kept its performance with orthosis patients, and although the graphs are similar, the standard deviation in most cases with the orthosis is slightly lower. 

\section{Conclusion}

We obtained promising results on the assessment of hand movement for occupational therapy using computer vision. The results obtained with the hand pose estimation are adequate and for simple movement sequences the angles computed are close to the real measurements. The system also works in patients with Rheumatoid Arthritis and with orthosis.

In the context of occupational therapy, the use of computer vision is less intrusive than goniometry. However, the precision of range of motion measurements needs further study and comparison.

Also, since the analysis of the skeleton is made in a posterior step, we can easily take advantage of recent advances in 3D hand pose estimation.

\ifFGfinal

\fi

\bibliographystyle{IEEEtran}
\bibliography{paper}

\begin{thebibliography}{10}
\providecommand{\url}[1]{#1}
\csname url@samestyle\endcsname
\providecommand{\newblock}{\relax}
\providecommand{\bibinfo}[2]{#2}
\providecommand{\BIBentrySTDinterwordspacing}{\spaceskip=0pt\relax}
\providecommand{\BIBentryALTinterwordstretchfactor}{4}
\providecommand{\BIBentryALTinterwordspacing}{\spaceskip=\fontdimen2\font plus
\BIBentryALTinterwordstretchfactor\fontdimen3\font minus
  \fontdimen4\font\relax}
\providecommand{\BIBforeignlanguage}[2]{{%
\expandafter\ifx\csname l@#1\endcsname\relax
\typeout{** WARNING: IEEEtran.bst: No hyphenation pattern has been}%
\typeout{** loaded for the language `#1'. Using the pattern for}%
\typeout{** the default language instead.}%
\else
\language=\csname l@#1\endcsname
\fi
#2}}
\providecommand{\BIBdecl}{\relax}
\BIBdecl

\bibitem{chen2017}
X.~Chen, G.~Wang, H.~Guo, and C.~Zhang, ``Pose guided structured region
  ensemble network for cascaded hand pose estimation,'' \emph{Neurocomputing},
  2018.

\bibitem{erol}
A.~Erol, G.~Bebis, M.~Nicolescu, R.~D. Boyle, and X.~Twombly, ``Vision-based
  hand pose estimation: A review,'' \emph{Computer Vision and Image
  Understanding}, vol. 108, no.~1, pp. 52--73, 2007.

\bibitem{wei2016convolutional}
S.-E. Wei, V.~Ramakrishna, T.~Kanade, and Y.~Sheikh, ``Convolutional pose
  machines,'' in \emph{Proceedings of the IEEE Conference on Computer Vision
  and Pattern Recognition}, 2016, pp. 4724--4732.

\bibitem{newell2016stacked}
A.~Newell, K.~Yang, and J.~Deng, ``Stacked hourglass networks for human pose
  estimation,'' in \emph{European Conference on Computer Vision}.\hskip 1em
  plus 0.5em minus 0.4em\relax Springer, 2016, pp. 483--499.

\bibitem{oberweger2017}
M.~Oberweger and V.~Lepetit, ``Deepprior++: Improving fast and accurate 3d hand
  pose estimation,'' in \emph{ICCV workshop}, vol. 840, 2017, p.~2.

\bibitem{oberweger}
M.~Oberweger, P.~Wohlhart, and V.~Lepetit, ``Hands deep in deep learning for
  hand pose estimation,'' in \emph{Computer Vision Winter Workshop}.\hskip 1em
  plus 0.5em minus 0.4em\relax ., 2015.

\bibitem{zhou}
Y.~Zhou, G.~Jiang, and Y.~Lin, ``A novel finger and hand pose estimation
  technique for real-time hand gesture recognition,'' \emph{Pattern
  Recognition}, vol.~49, pp. 102--114, 2016.

\bibitem{ge2017}
L.~Ge, H.~Liang, J.~Yuan, and D.~Thalmann, ``{3D} convolutional neural networks
  for efficient and robust hand pose estimation from single depth images,'' in
  \emph{Proc. CVPR}, 2017.

\bibitem{guo2017}
H.~Guo, G.~Wang, X.~Chen, C.~Zhang, F.~Qiao, and H.~Yang, ``Region ensemble
  network: Improving convolutional network for hand pose estimation,'' in
  \emph{Image Processing (ICIP), 2017 IEEE International Conference on}.\hskip
  1em plus 0.5em minus 0.4em\relax IEEE, 2017, pp. 4512--4516.

\bibitem{wan17}
C.~Wan, T.~Probst, L.~Van~Gool, and A.~Yao, ``Crossing nets: Combining gans and
  vaes with a shared latent space for hand pose estimation,'' in \emph{2017
  IEEE Conference on Computer Vision and Pattern Recognition (CVPR)}.\hskip 1em
  plus 0.5em minus 0.4em\relax IEEE, 2017.

\bibitem{yuan2017}
S.~Yuan, Q.~Ye, B.~Stenger, S.~Jain, and T.-K. Kim, ``Bighand2. 2m benchmark:
  Hand pose dataset and state of the art analysis,'' in \emph{Computer Vision
  and Pattern Recognition (CVPR), 2017 IEEE Conference on}.\hskip 1em plus
  0.5em minus 0.4em\relax IEEE, 2017, pp. 2605--2613.

\bibitem{garcia2017}
G.~Garcia-Hernando, S.~Yuan, S.~Baek, and T.-K. Kim, ``First-person hand action
  benchmark with rgb-d videos and 3d hand pose annotations,'' in \emph{Computer
  Vision and Pattern Recognition (CVPR), 2018 IEEE Conference on}, vol.~1,
  no.~2, 2018.

\bibitem{handsmillion}
\BIBentryALTinterwordspacing
S.~Yuan, Q.~Ye, G.~Garcia{-}Hernando, and T.~Kim, ``The 2017 hands in the
  million challenge on 3d hand pose estimation,'' \emph{CoRR}, vol.
  abs/1707.02237, 2017. [Online]. Available:
  \url{http://arxiv.org/abs/1707.02237}
\BIBentrySTDinterwordspacing

\bibitem{yuan20173d}
S.~Yuan, G.~Garcia-Hernando, B.~Stenger, G.~Moon, J.~Yong~Chang, K.~Mu~Lee,
  P.~Molchanov, J.~Kautz, S.~Honari, L.~Ge \emph{et~al.}, ``Depth-based {td3D}
  hand pose estimation: From current achievements to future goals,'' in
  \emph{The IEEE Conference on Computer Vision and Pattern Recognition (CVPR)},
  2018.

\bibitem{zimmermann}
C.~Zimmermann and T.~Brox, ``Learning to estimate 3d hand pose from single rgb
  images,'' in \emph{International Conference on Computer Vision}, vol.~1,
  no.~2, 2017, p.~3.

\bibitem{panteleris}
P.~Panteleris and A.~Argyros, ``Back to rgb: 3d tracking of hands and
  hand-object interactions based on short-baseline stereo,'' in
  \emph{Proceedings of the IEEE International Conference on Computer Vision},
  2017, pp. 575--584.

\bibitem{mueller2017ganerated}
\BIBentryALTinterwordspacing
F.~Mueller, F.~Bernard, O.~Sotnychenko, D.~Mehta, S.~Sridhar, D.~Casas, and
  C.~Theobalt, ``Ganerated hands for real-time 3d hand tracking from monocular
  rgb,'' in \emph{Proceedings of Computer Vision and Pattern Recognition
  ({CVPR})}, June 2018. [Online]. Available:
  \url{https://handtracker.mpi-inf.mpg.de/projects/GANeratedHands/}
\BIBentrySTDinterwordspacing

\bibitem{panteleris2017using}
P.~Panteleris, I.~Oikonomidis, and A.~Argyros, ``Using a single rgb frame for
  real time 3d hand pose estimation in the wild,'' in \emph{Applications of
  Computer Vision (WACV), 2018 IEEE Winter Conference on}.\hskip 1em plus 0.5em
  minus 0.4em\relax IEEE, 2018, pp. 436--445.

\bibitem{ellis1997joint}
B.~Ellis, A.~Bruton, and J.~R. Goddard, ``Joint angle measurement: a
  comparative study of the reliability of goniometry and wire tracing for the
  hand,'' \emph{Clinical rehabilitation}, vol.~11, no.~4, pp. 314--320, 1997.

\bibitem{bruton1999comparison}
A.~Bruton, B.~Ellis, and J.~Goddard, ``Comparison of visual estimation and
  goniometry for assessment of metacarpophalangeal joint angle,''
  \emph{Physiotherapy}, vol.~85, no.~4, pp. 201--208, 1999.

\bibitem{carvalho2012analise}
R.~M. F.~d. Carvalho, N.~Mazzer, C.~H. Barbieri \emph{et~al.}, ``An{\'a}lise da
  confiabilidade e reprodutibilidade da goniometria em rela{\c{c}}{\~a}o {\`a}
  fotogrametria na m{\~a}o,'' \emph{Acta Ortop{\'e}dica Brasileira}, vol.~20,
  no.~3, pp. 139--149, 2012.

\bibitem{tajali2016reliability}
S.~B. Tajali, J.~C. MacDermid, R.~Grewal, and C.~Young, ``Reliability and
  validity of electro-goniometric range of motion measurements in patients with
  hand and wrist limitations,'' \emph{The open orthopaedics journal}, vol.~10,
  p. 190, 2016.

\bibitem{pereira2017reliability}
L.~C. Pereira, S.~Rwakabayiza, E.~L{\'e}cureux, and B.~M. Jolles, ``Reliability
  of the knee smartphone-application goniometer in the acute orthopedic
  setting,'' \emph{The journal of knee surgery}, vol.~30, no.~03, pp. 223--230,
  2017.

\bibitem{limafisiomotion}
L.~Lima, J.~Melo, T.~Fragoso, T.~Vieira, and M.~Oliveira, ``Fisiomotion:
  Sistema de avalia{\c{c}}{\~a}o de pacientes portadores de artrite reumatoide
  usando sensor de movimentos,'' in \emph{XXV Congresso Brasileiro de
  Engenharia Biomédica - CBEB 2016}, 2016.

\bibitem{GoiaOrthosis}
\BIBentryALTinterwordspacing
D.~N. Goia, C.~A. Fortulan, B.~M. Purquerio, and V.~M.~C. Elui,
  ``\BIBforeignlanguage{en}{A new concept of orthosis for correcting fingers
  ulnar deviation},'' \emph{\BIBforeignlanguage{en}{Research on Biomedical
  Engineering}}, vol.~33, pp. 50--57, 03 2017. [Online]. Available:
  \url{http://www.scielo.br/scielo.php?script=sci_arttext&pid=S2446-47402017000100050&nrm=iso}
\BIBentrySTDinterwordspacing

\end{thebibliography}
\end{document}